\documentclass[10pt]{article} 
\pdfoutput=1 
\usepackage[preprint]{tmlr}
\usepackage{rawfonts, amsmath, amsfonts, pgf, caption, subfig, makecell, lastpage}

\usepackage[hidelinks]{hyperref}
\usepackage{url}

\DeclareUnicodeCharacter{2212}{-}

\newcommand{\thickhline}{%
\noalign {\ifnum 0=`}\fi \hrule height 1pt
  \futurelet \reserved@a \@xhline
}

\title{Revisiting Gaussian Neurons for Online Clustering with Unknown Number of
  Clusters}

\author{\name Ole Christian Eidheim \email ole.c.eidheim@ntnu.no \\ \addr
  Norwegian University of Science and Technology,\\ Department of Computer
  Science,\\ NO-7491 Trondheim, Norway}

\begin{document}

\maketitle

\begin{abstract}%
  Despite the recent success of artificial neural networks, more biologically
  plausible learning methods may be needed to resolve the weaknesses of
  backpropagation trained models such as catastrophic forgetting and adversarial
  attacks. Although these weaknesses are not specifically addressed, a novel
  local learning rule is presented that performs online clustering with an upper
  limit on the number of clusters to be found rather than a fixed cluster count.
  Instead of using orthogonal weight or output activation constraints, activation
  sparsity is achieved by mutual repulsion of lateral Gaussian neurons ensuring
  that multiple neuron centers cannot occupy the same location in the input
  domain. An update method is also presented for adjusting the widths of the
  Gaussian neurons in cases where the data samples can be represented by means
  and variances. The algorithms were applied on the MNIST and CIFAR-10 datasets
  to create filters capturing the input patterns of pixel patches of various
  sizes. The experimental results\footnote{The source code to reproduce the figures in this article is available at
    \url{https://gitlab.com/eidheim/gaussian-neurons-for-online-clustering}}
  demonstrate stability in the learned parameters across a large number of
  training samples.
\end{abstract}

\section{Introduction}

Machine learning models trained through backpropagation have become widely
popular in the last decade since AlexNet \citep{krizhevsky2012imagenet}.
Backpropagation, however, is not considered biologically plausible
\citep{bengio2016biologically}, and the trained models are among other things
susceptible to catastrophic forgetting
\citep{mccloskey1989catastrophic,ratcliff1990connectionist} and adversarial
attacks \citep{szegedy2014intriguing}.

More biologically plausible methods would need to incorporate not only local
learning rules and unsupervised learning, but also have properties such as
sparser activations, capacity for distribution changes of the input, not prone
to catastrophic forgetting, and work in an online setting, where samples are
processed sequentially during learning one at a time.

One way to address distribution shift in online clustering is to utilize
overparameterized models that have additional parameters available to model
input from subsequent unknown distributions. This will require that the
parameters used to capture previous data be less likely to be adjusted, but
model susceptibility to catastrophic forgetting would consequently be
alleviated.

In this article, a novel learning rule is presented that utilizes neurons
expressed with Gaussian functions. For a given neuron, the online method has an
attraction term toward the current sample, and an inhibition term that ensures
reduced overlap between the Gaussian neurons in the same layer to achieve
activation sparsity. This makes it possible to have an overparameterized model
with more neurons in a layer than is needed to represent the input. Some
neurons will model previous input samples, while other neurons can adapt to new
input from a possibly different distribution. Although the inhibition term does
not fully resolve the possibility of catastrophic forgetting, specialized rules
may be added since the neurons representing already sampled data can be
identified.

Additionally, an update method for the neuron widths is presented, where the
sample variances are measured from the data samples, and used to perform a
constrained update of the neuron widths toward the sample variances. This can
in certain conditions lead to a more robust learning rule whose results are
less affected by the choice of initial neuron widths.

A limitation of the presented method is the use of isotropic Gaussian
functions, employing scalar variances instead of covariance matrices. Moreover,
Gaussian functions require additional computational resources and can be more
numerically unstable compared to linear functions. Gaussian functions can,
however, be approximated while retaining desired properties, for instance by
using piecewise linear functions.

Finally, no claims are made that the proposed learning rule achieves better
clustering results compared to previous work. On the other hand, the presented
methods have a few interesting properties, as demonstrated in Section
\ref{results} such as examples of robustness to model disruption and
distribution shift, which merit further studies toward more biologically
plausible artificial neural networks that may be less susceptible to the
weaknesses of current trained models.

\section{Related Work}

Clustering in an online setting has previously been studied extensively
\citep{du2010clustering}, including for instance MacQueen's \emph{k}-means
\citep{macqueen1967some}, self-organizing maps \citep{kohonen1982self},
adaptive resonance theory \citep{carpenter1987massively}, and neural gas
\citep{fritzke1995growing}. These algorithms typically implement a
winner-take-all scheme, where only the winning neuron is adjusted with respect
to a given sample.

Other methods extends Oja's learning rule \citep{oja1982simplified} to achieve
online clustering, such as White's competitive Hebbian learning rule
\citep{white1992competitive} that adds a lateral inhibition term to the Oja's
learning rule, and Decorrelated Hebbian Learning \citep{deco1995decorrelated}
where softmax normalized Gaussian functions are used for neuron activations.
More recently, \citet{pehlevan2014hebbian} presented an online clustering
algorithm using symmetric non-negative matrix factorization based on the works
of \citet{ding2005equivalence,kuang2012symmetric}, and
\citet{krotov2019unsupervised} extended the Oja's learning rule with, moreover,
a lateral inhibition term. The latter method was parallelized on GPUs in
\citet{grinberg2019local} and \citet{talloen2021pytorchhebbian}. All of these
methods, however, use a fixed number of clusters that the algorithms are
predetermined to find.

Current artificial neuron models are overly simplified compared to cortical
neurons \citep{beniaguev2021single}. In artificial neural networks, an
activation function is usually used with a linear combination of input
variables. On the other hand, clustering algorithms such as \emph{k}-means and
Gaussian mixture models use distance and multivariate Gaussian functions,
respectively. The centroids or means of Gaussian functions capture more
significantly both the angle and magnitude of the input vector, than that of
linear combinations of input variables. Furthermore, variance and especially
covariance provide additional flexibility for modeling the input. To model a
convex region, for example, an artificial neural network using linear
combination of the input variables will need two fully-connected layers each
with a non-linear activation function, typically Rectified Linear Units since
\citet{glorat2011relu}. The first layer will then represent hyperplanes in the
input domain, and the second layer may combine these hyperplanes to convex
regions. Gaussian functions, on the other hand, innately capture convex
regions, potentially reducing the number of layers required for a given task.
In terms of number of layers, a biological cortex cannot process signals
sequentially as many times as the deeper artificial feedforward networks do,
due to the relatively low firing rate in biological neurons
\citep{wang2016firing}. As a consequence, more biologically plausible models
must perhaps be more shallow than is common in today's deep artificial
networks.

Sparse representations can be an effective way to reduce the dimensionality of
the input space, in addition to having the potential for more efficient use of
resources. Methods relying on linear combinations of the input variables, such
as sparse coding \citep{olshausen1997sparse}, typically add a regularization
term on the output activations. These algorithms, however, depend on
reconstruction errors and are therefore not viable in certain unsupervised
settings such as clustering. Another possibility is to regularize the linear
coefficients directly as in competitive Hebbian learning
\citep{white1992competitive} or independent component analysis
\citep{comon1994independent}, but these methods rely on orthogonal constraints,
which limits the clustering method, i.e. the number of clusters found is
generally less than or equal to the input dimensions, and two or more clusters
cannot exist along a given vector from the origin. On the other hand,
\citet{deco1995decorrelated} made use of softmax normalized Gaussian neurons
and regularization was achieved by penalizing overlapping Gaussian functions in
the same layer. The neuron and cost function of this method, however, can
result in learned cluster centers that are outside of the input domain, and
these centers would then poorly represent the true cluster centers. Moreover,
the proposed learning rule does not require a softmax function to produce the
neuron outputs.

\section{Learning Rule}

\begin{figure}[tb]
  \begin{center}
    \resizebox{0.5\textwidth}{!}{
      \input{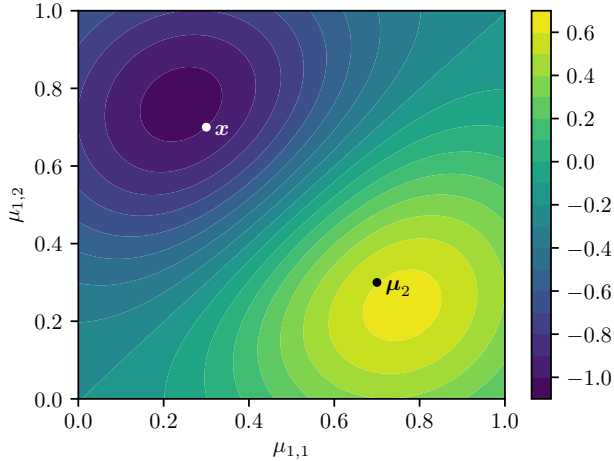}
    }
  \end{center}
  \caption{
    The function $F$ with $K=2$ Gaussian neurons parameterized by the first neuron
    $\boldsymbol{\mu}_1$, where $\boldsymbol{x}=\begin{pmatrix} 0.3 & 0.7
      \end{pmatrix}$, $\boldsymbol{\mu}_2
      = \begin{pmatrix} 0.7 & 0.3
      \end{pmatrix}$, $\sigma_1 = \sigma_2 = 0.2$, and $\lambda=\frac{1}{2}$. Minimizing $F$ with respect to
    $\boldsymbol{\mu}_1$ can be achieved by moving $\boldsymbol{\mu}_1$
    closer to $\boldsymbol{x}$ and farther from $\boldsymbol{\mu}_2$. The
    minimum of $F$ is here at
    $\boldsymbol{\mu}_1 \approx \begin{pmatrix} 0.24 & 0.76
      \end{pmatrix}$, although the presented learning rule will only make small changes to the
    cluster centers $\boldsymbol{\mu}_{1:K}$ for each $\boldsymbol{x}$.
    Furthermore,
    setting $\boldsymbol{\mu}_2 = \boldsymbol{x}$ would lead to a flat surface where
    $F = -1$, and in this case there would be no attraction or repulsion on
    $\boldsymbol{\mu}_1$.} \label{figure1}
\end{figure}

Let $\boldsymbol{x} \in \mathbb{R}^D$ be an input vector, and
$\boldsymbol{\mu}_i \in \mathbb{R}^D$ and $\sigma_i \in \mathbb{R}_{>0}$ be
center and width, respectively, of the $i$-th Gaussian neuron in a layer, where
$i \in \{1,\ldots,K\}$. The output of the $i$-th neuron is then defined as:

\[
  f_i(\boldsymbol{x}) = e^{-{\lVert \boldsymbol{x} -
  \boldsymbol{\mu}_i \rVert}_2^2/\sigma_i} \]

\noindent where ${\lVert \cdot \rVert}_2$ denotes the $l_2$-norm. The objective
of the learning rule is to minimize the cost function $E$ with respect to the
centers $\boldsymbol{\mu}_{1:K}$:

\[
  E(\mathbb{X};\boldsymbol{\mu}_1,\ldots,\boldsymbol{\mu}_K,\sigma_1,\ldots,\sigma_K)
  = \sum_{\boldsymbol{x} \in \mathbb{X}} \sum_{i=1}^K \left (
  -f_i(\boldsymbol{x}) + \lambda \sum_{j \ne i}
  f_j(\boldsymbol{\mu}_i) \right )
\]

\noindent where $\mathbb{X} \in \mathbb{R}^{N \times D}$ is a potentially very
large set of $N$ training examples that $\boldsymbol{x}$ is sampled from, $K$
is the number of Gaussian neurons, and $\lambda \in \mathbb{R}_{>0}$ controls
the level of inhibition.

The first term of $E$ attracts the cluster centers $\boldsymbol{\mu}_{1:K}$
toward the input $\boldsymbol{x}$, while the second term repels the different
cluster centers away from each other depending on the Gaussian locations and
widths. Interestingly, setting $\lambda \ge \frac{1}{2}$ would nullify all
attraction toward the given input if a cluster center already has the same
position as the input and the Gaussian widths are equal. Lowering $\lambda$ and
$\sigma_{1:K}$ has the same effect, that is more input patterns can be learned,
but smaller widths $\sigma_{1:K}$ may lead to insufficient attraction of the
cluster centers toward the input.

In order to minimize $E$ in an online setting, the learning rule will update
$\boldsymbol{\mu}_{1:K}$ by small steps in the opposite direction, and
proportionally to the magnitude, of the gradient of the second sum of $E$:

\[
  F(\boldsymbol{x};\boldsymbol{\mu}_1,\ldots,\boldsymbol{\mu}_K,\sigma_1,\ldots,\sigma_K)
  = \sum_{i=1}^K \left ( -f_i(\boldsymbol{x}) + \lambda
  \sum_{j \ne i} f_j(\boldsymbol{\mu}_i) \right ) \]

\noindent at input $\boldsymbol{x}$. See Figure \ref{figure1} for an example
surface of $F$ with two neurons. For each input $\boldsymbol{x}$, the cluster
centers will be updated as follows:

\[
  \boldsymbol{\mu}_i = \boldsymbol{\mu}_i + \Delta
  \boldsymbol{\mu}_i =
  \boldsymbol{\mu}_i - \frac{1}{2} \eta_\mu
  \frac{\partial F}{\partial
    \boldsymbol{\mu}_i}
\]

\noindent where $\eta_\mu \in \mathbb{R}_{>0}$ is the learning rate. The
learning rule then becomes:

\begin{equation}
  \label{learning rule}
  \Delta \boldsymbol{\mu}_i = \eta_\mu \left (
  \frac{f_i(\boldsymbol{x})}{\sigma_i}(\boldsymbol{x}-\boldsymbol{\mu}_i) -
  \lambda \sum_{j \ne i} \left (\frac{f_i(\boldsymbol{\mu}_j)}
      {\sigma_i} + \frac{f_j(\boldsymbol{\mu}_i)}{\sigma_j} \right )
    (\boldsymbol{\mu}_j - \boldsymbol{\mu}_i) \right )
\end{equation}

\noindent If $\sigma_i = \sigma \  \forall i \in \{1, \ldots, K\}$, the
learning rule can be simplified to:

\[ \Delta \boldsymbol{\mu}_i = \frac{\eta_\mu}{\sigma} \left (
  f_i(\boldsymbol{x})(\boldsymbol{x}-\boldsymbol{\mu}_i)
  - 2 \lambda \sum_{j \ne i} f_i(\boldsymbol{\mu}_j)
  (\boldsymbol{\mu}_j - \boldsymbol{\mu}_i) \right )
\]

\noindent When applying the learning rule in Equation \ref{learning rule}, some
cluster centers may be repelled outside of the input domain. However, this is a
beneficial property that can be utilized to identify learned cluster centers
that correspond to patterns found from the input, as these centers will lie
within or relatively close to the input domain.

The maximum distance a cluster center can be pushed outside of the input domain
is constrained by the input domain itself, the parameters of Equation
\ref{learning rule}, and the number of clusters $K$. The maximum distance goes
toward $\infty$ as the number of clusters approaches $\infty$, but considering
$D=1$, $K=2$, $\lambda=\frac{1}{2}$, $\mu_2=0$, $\sigma_1 = \sigma_2 = \sigma$,
and $f_i(x) \to 0 \ \forall i \in \{1, 2\} \ \forall x \in \mathbb{R}$, the
maximum distance between $\mu_1$ and $\mu_2$ converges to the solution of the
ordinary differential equation at $\mu_1(t)$ as $t \to \infty$:

\[
  \frac{\partial \mu_1}{\partial t} = -\frac{\partial F}{\partial \mu_1} =
  2 \frac{\mu_1}{\sigma} e^{-\mu_1^2/\sigma},\quad
  \mu_1(0) > 0 \]

\noindent Although in practice, the learning rate, the number of iterations,
and the precision of the data types will limit the above distance.

\subsection{Update Method for Neuron Widths}

Having sparsity constraints given by Gaussian functions enables adjustments of
the activation sparsity based on sample measures. The Gaussian widths
$\sigma_{1:K}$ can be derived from the data $\mathbb{X}$ if a data sample is
instead represented by a mean vector $\boldsymbol{x} = \boldsymbol{\mu}_x$ and
a width scalar $\sigma_x$. This addition to the learning rule alters a neuron
width $\sigma_i$ toward the given data sample width $\sigma_x$ depending on
constraints such as how well the $i$-th Gaussian matches the data sample.
Furthermore, care must be taken to avoid neuron functions collapsing into each
other, i.e. resulting in several tight clusters near similar input patterns
that have a relatively high probability of being sampled.

Let the measure of a sample be:

\[
  f_x(\boldsymbol{\mu}_i) = e^{-\lVert \boldsymbol{\mu}_i - \boldsymbol{\mu}_x
      \rVert _2^2/\sigma_x} \]

\noindent and the update of the neuron widths be defined as:

\begin{equation}
  \label{width update}
  \Delta \sigma_i = \eta_\sigma \max \left \{ f_x(\boldsymbol{\mu}_i) -
  2 \lambda \sum_{j \ne i} f_j(\boldsymbol{\mu}_i), 0 \right \}
  f_i(\boldsymbol{\mu}_x) (\sigma_x - \sigma_i)
\end{equation}

\noindent where $\eta_\sigma \in \mathbb{R}_{>0}$ controls the rate of change,
and $\lambda$ is the inhibition level used in Equation \ref{learning rule}. The
change of a neuron width $\sigma_i$ is thus relying on the neuron mean with
respect to the sample measure inhibited by the other lateral neurons, and the
sample mean with respect to the neuron function. The $max\{\cdot\}$ function
ensures that the change of $\sigma_i$ is always $0$ or toward $\sigma_x$.

Equations \ref{learning rule} and \ref{width update} can be applied
independently, but adjusting means and widths interchangeably can be
advantageous. For example, by enabling additional clusters to be found after
the already identified clusters have had their widths reduced. Experiments
employing both Equations \ref{learning rule} and \ref{width update} are
presented in Subsection \ref{Varying Neuron Widths}.

\section{Results and Discussion}
\label{results}

The presented learning rule was tested on image patches from the MNIST
\citep{lecun1998gradient} and CIFAR-10 \citep{krizhevsky2009learning} datasets.
The $D$ pixel patch intensities were scaled to be from $0$ to $1$. No other
preprocessing was applied to the images or image patches.

\subsection{MNIST}

\begin{figure}[tb]
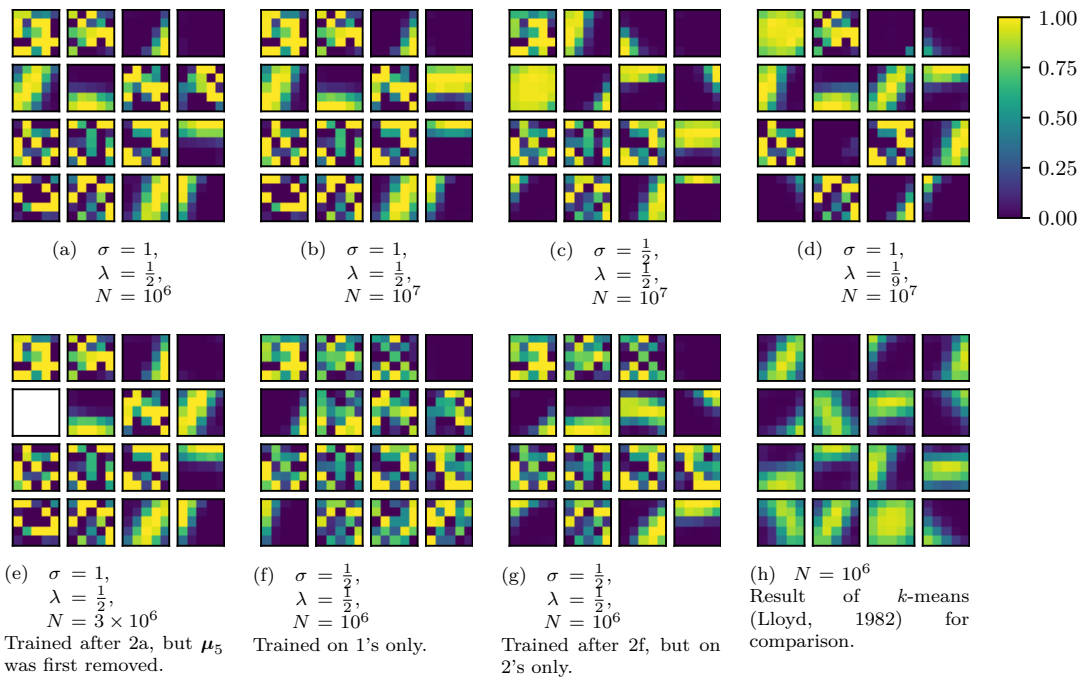

  \centering
  \resizebox{0.9\textwidth}{!}{
    \begin{tabular}{lllll}
      \subfloat[][\label{2a}{{\setlength\tabcolsep{1.5pt}
              \begin{tabular}[t]{ccl}
                $\sigma$ & $=$ & $1$, \\
                $\lambda$ & $=$ & $\frac{1}{2}$, \\
                $N$ & $=$ & $10^6$
              \end{tabular}}}]{\input{figure2a.pgf}}
      &
      \subfloat[][\label{2b}{{\setlength\tabcolsep{1.5pt}
              \begin{tabular}[t]{ccl}
                $\sigma$ & $=$ & $1$, \\
                $\lambda$ & $=$ & $\frac{1}{2}$, \\
                $N$ & $=$ & $10^7$
              \end{tabular}}}]{\input{figure2b.pgf}}
      &
      \subfloat[][\label{2c}{{\setlength\tabcolsep{1.5pt}
              \begin{tabular}[t]{ccl}
                $\sigma$ & $=$ & $\frac{1}{2}$, \\
                $\lambda$ & $=$ & $\frac{1}{2}$, \\
                $N$ & $=$ & $10^7$
              \end{tabular}}}]{\input{figure2c.pgf}}
      &
      \subfloat[][\label{2d}{{\setlength\tabcolsep{1.5pt}
              \begin{tabular}[t]{ccl}
                $\sigma$ & $=$ & $1$, \\
                $\lambda$ & $=$ & $\frac{1}{9}$, \\
                $N$ & $=$ & $10^7$
              \end{tabular}}}]{\input{figure2d.pgf}}
      & \input{figure2_colorbar.pgf}
      \\
      \vspace{0mm}
      \subfloat[][\label{2e}{{\setlength\tabcolsep{1.5pt}
              \begin{tabular}[t]{ccl}
                $\sigma$ & $=$ & $1$, \\
                $\lambda$ & $=$ & $\frac{1}{2}$, \\
                $N$ & $=$ & $3 \times 10^6$
              \end{tabular}
              \\\normalfont{Trained after \ref{2a}, but $\boldsymbol{\mu}_5$ was first removed.}}}]{\input{figure2e.pgf}} & 
      \subfloat[][\label{2f}{{\setlength\tabcolsep{1.5pt}
              \begin{tabular}[t]{ccl}
                $\sigma$ & $=$ & $\frac{1}{2}$, \\
                $\lambda$ & $=$ & $\frac{1}{2}$, \\
                $N$ & $=$ & $10^6$
              \end{tabular}
              \\Trained on 1's only.}}]{\input{figure2f.pgf}}
      &
      \subfloat[][\label{2g}{{\setlength\tabcolsep{1.5pt}
              \begin{tabular}[t]{ccl}
                $\sigma$ & $=$ & $\frac{1}{2}$, \\
                $\lambda$ & $=$ & $\frac{1}{2}$, \\
                $N$ & $=$ & $10^6$
              \end{tabular}
              \\Trained after \ref{2f}, but on 2's only.}}]{\input{figure2g.pgf}}
      &
      \subfloat[][\label{2h}{{\setlength\tabcolsep{1.5pt}
              \begin{tabular}[t]{ccl}
                $N$ & $=$ & $10^6$
              \end{tabular}
              \\Result of \emph{k}-means \citep{lloyd1982Least} for
              comparison.}}]{\input{figure2h.pgf}}
    \end{tabular} } \caption{The resulting
  centers or filters from applying the presented learning rule (\ref{2a} to
  \ref{2g}) with $K=16$ neurons, $N$ $5 \times 5$ pixel patches uniformly sampled
  from the MNIST training dataset, Gaussian widths $\sigma_i = \sigma \ \forall i
    \in \{1, \ldots, K\}$, inhibition level $\lambda$, and learning rate
  $\eta_\mu=10^{-1}$. The learned filters can for instance act as edge detectors,
  while the remaining filters were pushed outside of the input domain and had a
  high cosine similarity to the initial filter values prior to training. The
  filters are enumerated left to right, top to bottom.} \label{figure2}
\end{figure}

$N$ uniformly random $D=5 \times 5$ pixel patches were sampled from the
training set containing $6 \times 10^6$ images of handwritten digits. Figure
\ref{figure2} shows the resulting centers, typically named filters in this
context, of $K=16$ clusters with Gaussian widths $\sigma_i = \sigma \ \forall i
  \in \{1, \ldots, K\}$, inhibition level $\lambda$, and learning rate
$\eta_\mu=10^{-1}$.

Some cluster centers have not changed significantly from training, while other
filters represent learned filters that for instance can act as edge detection
filters. The unchanged filters have typically been pushed outside of the input
domain, and have a high cosine similarity to the initial randomized filters, as
shown in Table \ref{table1} for the filters depicted in Subfigure \ref{2a}.

\begin{table}[h]
  \centering
  \resizebox{\textwidth}{!}{
    \begin{tabular}{l|cccccccccccccccc}
      \Xhline{2\arrayrulewidth}
      \space &
      $\mu_1$ & $\mu_2$ & $\boldsymbol{\mu}_3$ & $\boldsymbol{\mu}_4$ &
      $\boldsymbol{\mu}_5$ & $\boldsymbol{\mu}_6$ & $\mu_7$ & $\mu_8$ & $\mu_9$ &
      $\mu_{10}$ & $\mu_{11}$ & $\boldsymbol{\mu}_{12}$ & $\mu_{13}$ & $\mu_{14}$ &
      $\boldsymbol{\mu}_{15}$ & $\boldsymbol{\mu}_{16}$
      \\ \hline
      $d_i$ & $1.5$ & $1.5$ & $\boldsymbol{1.0}$ & $\boldsymbol{1.0}$ &
      $\boldsymbol{0.8}$ & $\boldsymbol{0.9}$ & $1.4$ & $1.4$ & $1.5$ & $1.4$ & $1.5$
      & $\boldsymbol{0.9}$ & $1.6$ & $1.5$ & $\boldsymbol{0.8}$ & $\boldsymbol{0.9}$
      \\
      $\cos \theta_i$ & $0.9$ & $0.8$ & $\boldsymbol{0.4}$ & $\boldsymbol{-0.1}$ &
      $\boldsymbol{0.8}$ & $\boldsymbol{0.7}$ & $0.9$ & $0.9$ & $0.9$ & $0.9$ & $0.9$
      & $\boldsymbol{0.7}$ & $0.9$ & $0.9$ & $\boldsymbol{0.8}$ & $\boldsymbol{0.6}$
      \\ \Xhline{2\arrayrulewidth}
    \end{tabular}
  } \caption{Scalar descriptors of the filters shown in
  Subfigure \ref{2a}, where $d_i={\lVert \boldsymbol{\mu}_i - \frac{1}{2} \rVert}_2 /
    \sqrt{\frac{D}{4}}$, $\sqrt{\frac{D}{4}}$ is the maximum distance
  from a point in the input domain to the input domain centroid, $\cos
    \theta_i=\boldsymbol{\mu}_i \cdot \boldsymbol{\mu}_i^{(init)} /
    ({\lVert
        \boldsymbol{\mu}_i \rVert}_2 {\lVert
    \boldsymbol{\mu}_i^{(init)} \rVert}_2)$,
  and $\boldsymbol{\mu}_i^{(init)}$ is the $i$-th initial randomized filter
  before training. The learned filters, which for instance can act as edge
  detectors, are marked in bold.} \label{table1}
\end{table}

The difference between the results shown in Subfigures \ref{2a} and \ref{2b} is
that the latter experiment was run on $10$ times as many data samples.
Regardless, the filters shown in Subfigure \ref{2b} are similar to those in
Subfigure \ref{2a}, demonstrating that the filters can be stable throughout a
high number of iterations. Both experiments used the same random generator
seed.

Additional patterns were found in the results shown in Subfigures \ref{2c} and
\ref{2d} by lowering either the Gaussian width $\sigma$ or the inhibition level
$\lambda$, respectively. $11$ and $12$ learned filters are present in
Subfigures \ref{2c} and \ref{2d} compared to $8$ filters in Subfigure \ref{2b},
with the same number of samples processed. Reducing the Gaussian widths can,
however, lead to fewer identified patterns since the pull toward some data
samples may be insufficient. In addition, if the inhibition level is too low,
filters can collapse into each other, forming several identical Gaussian
centers. For example, 2 equal Gaussian centers $\boldsymbol{\mu}_4 =
  \boldsymbol{\mu}_{13} \approx \begin{pmatrix} 0, \ldots, 0
  \end{pmatrix}$ were found when the
experiment shown in Subfigure \ref{2d} was run with the inhibition level
lowered to $\lambda=10^{-1}$.

In the result shown in Subfigure \ref{2e}, the learned filters remain
approximately the same as in Subfigure \ref{2a}, despite the fact that the
learned filter $\boldsymbol{\mu}_5$ was removed prior to training on additional
$N=3 \times 10^6$ data samples. The pattern learned by filter
$\boldsymbol{\mu}_5$ in Subfigure \ref{2a} was then relearned by filter
$\boldsymbol{\mu}_8$ in Subfigure \ref{2e}. This result demonstrates potential
stability of the learning rule as a pattern can be relearned after disruption
while the other learned filters can remain largely unaltered.

Subfigure \ref{2f} shows the result of training solely on images depicting the
number 1, and this model was subsequently used as a starting point to produce
the result shown in Subfigure \ref{2g}, where only images corresponding to the
number 2 were used during training. Even though there was a distribution change
of the inputs, the three filters found in Subfigure \ref{2f} are largely
similar to the corresponding filters in Subfigure \ref{2g}, while additional
filters were learned as well. The additional filters mainly represents patterns
found exclusively in the latter dataset.

For comparison, Subfigure \ref{2h} shows the resulting filters from the
\emph{k}-means algorithm \citep{lloyd1982Least} with the cluster count set to
$16$. The proposed learning rule will most likely not learn as many filters
from the MNIST dataset given the relatively small pixel patch size without
setting the Gaussian width $\sigma$ or inhibition level $\lambda$ too low.

\subsubsection{Larger Pixel Patches}

\begin{figure}[tb]
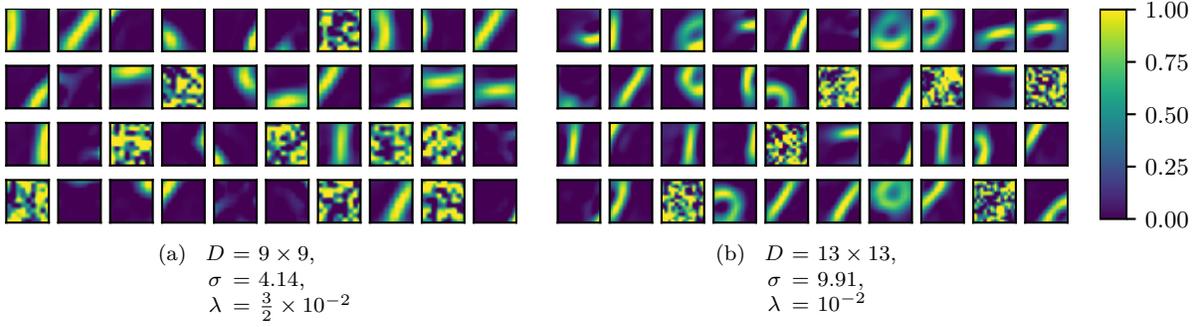

  \centering
  \resizebox{\textwidth}{!}{
    \begin{tabular}{lll}
      \subfloat[][\label{3a}{{\setlength\tabcolsep{1.5pt}
              \begin{tabular}[t]{ccl}
                $D$ & $=$ & $9 \times 9$, \\
                $\sigma$ & $=$ & $4.14$, \\
                $\lambda$ & $=$ & $\frac{3}{2} \times 10^{-2}$
              \end{tabular}}}]{\input{figure3a.pgf}}
      &
      \subfloat[][\label{3b}{{\setlength\tabcolsep{1.5pt}
              \begin{tabular}[t]{ccl}
                $D$ & $=$ & $13 \times 13$, \\
                $\sigma$ & $=$ & $9.91$, \\
                $\lambda$ & $=$ & $10^{-2}$
              \end{tabular}}}]{\input{figure3b.pgf}}
      &
      \input{figure3_colorbar.pgf}
    \end{tabular}
  }
  \caption{The results of employing the learning rule on larger image patches. In
  both experiments, the number of samples was $N=10^7$, the Gaussian widths set
  to $\sigma_i = \sigma \ \forall i \in \{1, \ldots, K\}$, and the learning rate
  equal to $\eta_\mu=10^{-1}$. More rounded filters were learned compared to
  those in Figure \ref{figure2}, but the Gaussian widths had to be increased in
  order to better capture the patterns in the larger input
  domain.}\label{figure3}
\end{figure}

Experiment results with $N=10^7$ samples of $D=9 \times 9$ and $D=13 \times 13$
pixel patches are shown in Figure \ref{figure3}. Similar to Figure
\ref{figure2}, the Gaussian widths were set to $\sigma_i = \sigma \ \forall i
  \in \{1, \ldots, K\}$. More rounded filters were learned, especially those
shown in Subfigure \ref{3b}, which can be used to detect parts of particularly
the handwritten digits $0$, $2$, $3$, $5$, $6$, $8$, and $9$.

As the number of input dimensions was increased, leading to a larger input
domain, the inhibition level was decreased and the Gaussian widths were
increased to better capture the patterns of the input samples. To make the
results more comparable with, for instance, the results in Subfigure \ref{2d},
the width $\sigma$ was calculated such that two neurons placed on opposite
positions of the input domains had equal repulsion, measured by the $l_2$-norm
normalized by the number of dimensions, independent of input samples:

\begin{equation}
  \begin{array}{lcr}
    \begin{aligned}
      \frac{1}{D}{\lVert \Delta \boldsymbol{\mu}_1 \rVert}_2
                                                              & =
      \frac{1}{\tilde{D}}{\lVert	\Delta \tilde{\boldsymbol{\mu}}_1
        \rVert}_2
      \\
      \frac{1}{D}{\left \lVert
      \frac{\lambda \boldsymbol{\mu}_1}{\sigma^2} e^{-{\lVert
      \boldsymbol{\mu}_1 \rVert}_2^2/\sigma} \right \rVert}_2 & =
      \frac{1}{\tilde{D}}{\left \lVert
      \frac{\tilde{\lambda} \tilde{\boldsymbol{\mu}}_1}{\tilde{\sigma}^2}
      e^{-{\lVert \tilde{\boldsymbol{\mu}}_1 \rVert}_2^2/\tilde{\sigma}} \right
      \rVert}_2
      \\
      \frac{\lambda}{\sqrt{D} \sigma^2} e^{-D/\sigma}         & =
      \frac{\tilde{\lambda}}{\sqrt{\tilde{D}} \tilde{\sigma}^2}
      e^{-\tilde{D} / \tilde{\sigma}}
    \end{aligned}
                  &
    \textrm{s.t.} &
    \begin{aligned}
      \sigma        & >
      0
      \\
      \frac{\partial}{\partial \sigma} \frac{\lambda}{\sqrt{D} \sigma^2}
      e^{-D/\sigma} & > 0
    \end{aligned}
  \end{array} \label{width equation}
\end{equation}

\noindent where $K=2$, $\boldsymbol{\mu}_1=(1,\ldots, 1)$,
$\boldsymbol{\mu}_2=(0,\ldots, 0)$, $\boldsymbol{\mu}_i \in
  \mathbb{R}^D$,
$f_1(\boldsymbol{x}) \to 0 \ \forall \boldsymbol{x} \in
  \mathbb{R}^D$,
$\tilde{\boldsymbol{\mu}}_1=(1,\ldots, 1)$,
$\tilde{\boldsymbol{\mu}}_2=(0,\ldots, 0)$, $\tilde{\boldsymbol{\mu}}_i \in
  \mathbb{R}^{\tilde{D}}$, and $\tilde{f}_1(\tilde{\boldsymbol{x}}) \to 0
  \ \forall \tilde{\boldsymbol{x}} \in
  \mathbb{R}^{\tilde{D}}$. The tilde symbol denotes the vectors and scalars used
in this case to produce the result presented in Subfigure \ref{2d}, and the
learning rate $\eta_\sigma$ was equal in both $\Delta \boldsymbol{\mu}_1$ and
$\Delta \tilde{\boldsymbol{\mu}}_1$.

From Equation \ref{width equation} then with $\tilde{D} = 5 \times 5$,
$\tilde{\sigma} = 1$, $\tilde{\lambda} = \frac{1}{9}$, the neuron widths for
computing the results shown in the Subfigures \ref{3a} and \ref{3b} were
approximated to be $\sigma = 4.14$ and $\sigma = 9.91$, respectively. The
inhibition level was made slightly different in order to find roughly the same
number of filters, where $\lambda=\frac{3}{2} \times 10^{-2}$ was used to
process $D=9 \times 9$ pixel patches and $\lambda=10^{-2}$ to process $D=13 \times 13$ pixel patches.

\subsection{CIFAR-10}

\begin{figure}[tb]
  \centering
  \resizebox{0.44\textwidth}{!}{
    \input{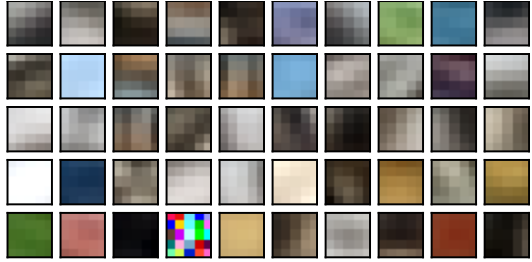}
  }
  \caption{The resulting filters from training on $D=5 \times 5 \times 3$ pixel
  patches from the CIFAR-10 dataset containing natural images, with $N=10^7$
  number of samples, Gaussian widths $\sigma_i = 1.75 \ \forall i \in \{1,
    \ldots, K\}$, inhibition level $\lambda=10^{-2}$, and learning rate
  $\eta_\mu=10^{-1}$. Some of the edge detecting filters are color independent,
  while some filters can be used to detect specific color patches such as sky and
  grass.}\label{figure4}
\end{figure}

The learning rule was also applied on the CIFAR-10 dataset
\citep{krizhevsky2009learning}, where $N$ uniformly random $D=5 \times 5 \times 3$ pixel patches were sampled from the training set containing $6 \times 10^6$
natural images, evenly portraying airplanes, cars, birds, cats, deer, dogs,
frogs, horses, ships, and trucks. Figure \ref{figure4} shows the resulting
filters of $K=50$ clusters with Gaussian widths $\sigma_i = \sigma = 1.75 \
  \forall i \in \{1, \ldots, K\}$, inhibition level $\lambda = 10^{-2}$, and
learning rate $\eta_\mu=10^{-1}$.

Some of the learned filters can be used to detect specific colored regions such
as sky and grass, while other filters can act as edge detectors. Some of the
edge detecting filters are gray and thus color independent. All filters except
one learned patterns from the pixel patch samples.

\subsection{Neuron Width Update}
\label{Varying Neuron Widths}

\begin{figure}[tb]
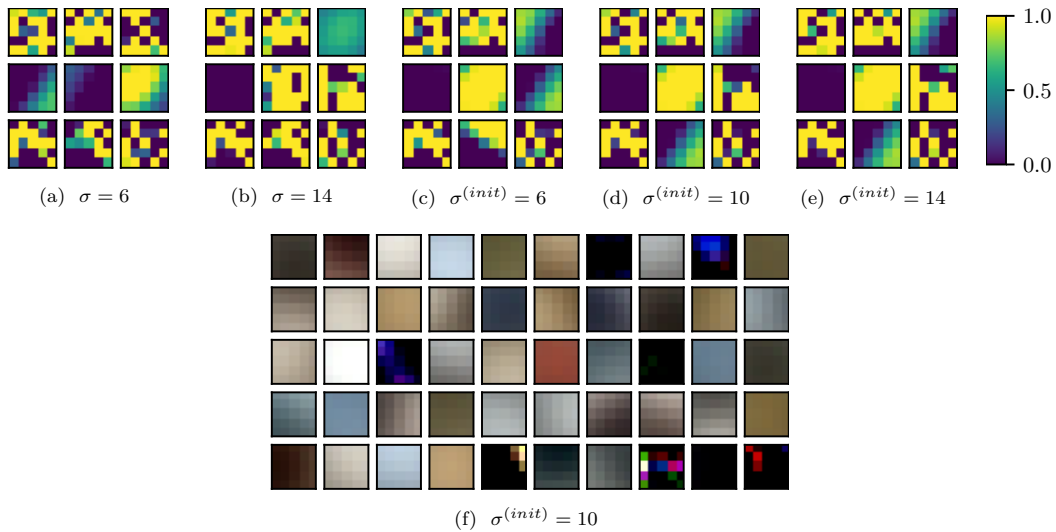

  \centering
  \resizebox{0.9\textwidth}{!}{
    \begin{tabular}{cc}
      \begin{tabular}{llllll}
        \subfloat[][\label{5a} $\sigma=6$]{\input{figure5a.pgf}} &
        \subfloat[][\label{5b} $\sigma=14$]{\input{figure5b.pgf}} &
        \subfloat[][\label{5c}
          $\sigma^{(init)}=6$]{\input{figure5c.pgf}} &
        \subfloat[][\label{5d}
          $\sigma^{(init)}=10$]{\input{figure5d.pgf}} &
        \subfloat[][\label{5e}
          $\sigma^{(init)}=14$]{\input{figure5e.pgf}} & \input{figure5_colorbar.pgf}
      \end{tabular} \\ \vspace{0mm} \subfloat[][\label{5f}
        $\sigma^{(init)}=10$]{\input{figure5f.pgf}}
    \end{tabular}}	\caption{The results of employing both Equation
  \ref{learning rule} and \ref{width update} on data sample measures from the
  MNIST and CIFAR-10 datasets. Subfigures \ref{5c} to \ref{5e} show similar
  results even though the initial neuron widths were different. On the other hand
  3 and 2 filters were learned when the neuron widths were fixed to $\sigma=6$
  and $\sigma=14$ as shown in Subfigures \ref{5a} and \ref{5b}, respectively. The
  result from Figure \ref{figure4} was reproduced in Subfigure \ref{5f} although
  with varying Gaussian widths. All results were produced using $N=10^7$ data
  samples and learning rates set to $\eta_\mu=\eta_\sigma=10^{-1}$.}
  \label{figure5}
\end{figure}

Both Equations \ref{learning rule} and \ref{width update} were employed to
produce the results shown in Figure \ref{figure5}, using $N=10^7$ data samples
and learning rate set to $\eta_\mu=\eta_\sigma=10^{-1}$. In Subfigures \ref{5c}
to \ref{5f}, the data sample mean
$\boldsymbol{x}=\boldsymbol{\mu}_x=\frac{1}{9}\sum_{i=1}^9
  \boldsymbol{s}_i$
and width $\sigma_x = 2 \textrm{Var}[\boldsymbol{s}_{1:9}] =
  \frac{2}{9}\sum_{i=1}^9(\boldsymbol{s}_i-\boldsymbol{\mu}_x)^2$ were measured
from $9$ neighboring $5 \times 5$ image patches $\boldsymbol{s}_{1:9}$ after
adding normal distributed noise with $0$ mean, and $0.1$ (Subfigures \ref{5c}
to \ref{5e}) or $0.15$ (Subfigure \ref{5f}) standard deviation, to each pixel
value. Neighboring image patches can share similar patterns, and noise was
added as a constraint to keep the Gaussian widths from collapsing toward $0$,
for example, when sampling homogeneous image regions.

The neuron means $\boldsymbol{\mu}_{1:K}$ and widths $\sigma_{1:K}$ were
updated in alternate succession, that is for each data sample, the neuron means
were first updated using Equation \ref{learning rule}, and then Equation
\ref{width update} was used to update the neuron widths.

Subfigures \ref{5c} to \ref{5e} show similar $K=9$ filters even though the
initial Gaussian widths $\sigma_i^{(init)}=\sigma^{(init)} \  \forall i \in
  \{1, \ldots, K\}$ were set to $\sigma^{(init)}=6$, $\sigma^{(init)}=10$, and
$\sigma^{(init)}=14$, respectively. For comparison, Subfigures \ref{5a} and
\ref{5b} show the resulting filters without neuron width updates where the
Gaussian widths $\sigma_i = \sigma \ \forall i \in \{1, \ldots, K\}$ were set
to, in the same order, $\sigma = 4$ and $\sigma = 14$. The inhibition level was
$\lambda = \frac{1}{2}$ for all Subfigures \ref{5a} to \ref{5e}.

The final Gaussian widths of the filters shown in Subfigure \ref{5d} were:

\[
  \sigma_{1:K}
  =
  \begin{pmatrix}
    10, 9.5, \boldsymbol{4.6}, \boldsymbol{0.6}, \boldsymbol{5}, 10, 10,
    \boldsymbol{4.6}, 10
  \end{pmatrix} \]

\noindent and the learned filters can in this case be
distinguished from the filters that have not learned any patterns by comparing
the final widths to the initial widths of the Gaussians.

Lastly, Subfigure \ref{5f} shows the result of applying the learning rule and
width update method on data samples from the CIFAR-10 dataset with initial
Gaussian width $\sigma^{(init)}=10$ and inhibition level $\lambda = 10^{-2}$.
Almost the same number of filters were learned compared to those learned in
Figure \ref{figure4}, although the initial widths were larger and each data
sample was extracted from $9$ neighboring $5 \times 5 \times 3$ pixel patches.
By using Equation \ref{width update}, however, the Gaussian widths of the
learned filters were reduced and thus made room for additional Gaussians within
the data domain.

\section{Conclusion and Future Work}

A novel unsupervised learning rule based on Gaussian functions is presented
that can perform online clustering without needing to specify the number of
clusters prior to training. The local learning rule is arguably more
biologically plausible compared to model optimization through backpropagation,
and the results demonstrate stability in the learned parameters during
training. Furthermore, an update method for the widths of the Gaussian
functions is presented, which can reduce the dependence on finely tuned
hyperparameters.

Opportunities for future work include investigations of the presented methods
in the following directions: (1) improve the computation speed through CPU or
GPU parallelization, (2) use anisotropic Gaussian functions, (3) train with
input data other than image patches, (4) layer-wise optimization of deeper
model architectures, (5) compare trained model susceptibility to adversarial
attacks against models optimized through backpropagation, (6) sample mean and
variance from subsequent video frames instead of neighboring image patches, and
(7) repurposing the Gaussian functions as Gaussian dendrites, and linearly
combine similar dendritic functions into artificial neurons to achieve greater
representational power of the artificial neurons.

\bibliography{article}
\bibliographystyle{tmlr}

\end{document}